\documentclass[journal]{IEEEtran}
\usepackage[utf8]{inputenc}
\usepackage{cite}

\usepackage{svg}
\usepackage{amsmath}
\usepackage{amssymb}
\usepackage{physics}
\usepackage{hyperref}
\usepackage{relsize}%
\usepackage{graphicx}
\usepackage{subfig}
\usepackage{xcolor}
\usepackage[utf8]{inputenc}
\usepackage[english]{babel}
\usepackage{amsthm}
\usepackage{xurl}

\usepackage{algorithmicx}
\usepackage[ruled,vlined, linesnumbered]{algorithm2e}
\usepackage{comment} 
\newcommand{\ignore}[1]{}
\usepackage{color} 
\usepackage{multirow}
\usepackage[T1]{fontenc}

\usepackage{algpseudocode}
\usepackage{bm}
\usepackage[font=small]{caption}
\usepackage{enumitem}

\makeatletter
\newcommand*{\rom}[1]{\expandafter\@slowromancap\romannumeral #1@}

\newcommand{\removelatexerror}{\let\@latex@error\@gobble}
\newcommand*{\Scale}[2][4]{\scalebox{#1}{$#2$}}%
\let\oldnl\nl
\newcommand{\nonl}{\renewcommand{\nl}{\let\nl\oldnl}}
\makeatother

\IEEEoverridecommandlockouts
\begin{document}

\title{\LARGE \bf Back-stepping Experience Replay with Application to Model-free Reinforcement Learning for a Soft Snake Robot}

\author{
Xinda Qi$^{1}$, 
Dong Chen$^{1,*}$, \textit{Member, IEEE}, Zhaojian Li$^{{2}}$,  \textit{Senior Member, IEEE},  Xiaobo Tan$^{1}$, \textit{Fellow, IEEE}
\vspace{-15pt}

\thanks{This work has been submitted to the IEEE for possible publication. Copyright may be transferred without notice, after which this version may no longer be accessible. This work was supported by the National Science Foundation (Grant CNS 2125484).}
\thanks{$^1$Xinda Qi, Dong Chen, and Xiaobo Tan are with the Department of Electrical and Computer Engineering, Michigan State University, Lansing, MI, 48824, USA. Email:{\tt\{qixinda, chendon9, xbtan\}@msu.edu}.}
\thanks{$^2$Zhaojian Li is with the Department of Mechanical Engineering, Michigan State University, Lansing, MI, 48824, USA. Email:{\tt\ lizhaoj1@msu.edu}.}
\thanks{$^*$ Corresponding author.}
}

\maketitle

\begin{abstract}
In this paper, we propose a novel technique, Back-stepping Experience Replay (BER), that is compatible with arbitrary off-policy reinforcement learning (RL) algorithms. BER aims to enhance learning efficiency in systems with approximate reversibility, reducing the need for complex reward shaping. The method constructs reversed trajectories using back-stepping transitions to reach random or fixed targets. Interpretable as a bi-directional approach, BER addresses inaccuracies in back-stepping transitions through a distillation of the replay experience during learning. Given the intricate nature of soft robots and their complex interactions with environments, we present an application of BER in a model-free RL approach for the locomotion and navigation of a soft snake robot, which is capable of serpentine motion enabled by anisotropic friction between the body and ground. In addition, a dynamic simulator is developed to assess the effectiveness and efficiency of the BER algorithm, in which the robot demonstrates successful learning (reaching a 100\% success rate) and adeptly reaches random targets, achieving an average speed 48\% faster than that of the best baseline approach.


\begin{IEEEkeywords}
Deep reinforcement learning, experience replay, soft robot, snake robot, locomotion, navigation.
\end{IEEEkeywords}
\end{abstract}

\IEEEpeerreviewmaketitle

\section{Introduction}\label{sec:1}

As a promising decision-making approach, reinforcement learning (RL) has drawn increasing attention for its ability to solve complex control problems and achieve generalization in both virtual and physical tasks, as evidenced in various applications, such as chess games \cite{silver2016mastering_long}, quadrupedal locomotion \cite{liu2020learning}, and autonomous driving \cite{chen2020autonomous, hua2023multi}. Considering the inherent infinite degrees of freedom of soft robots and their complicated interactions with environments \cite{lee2017soft}, RL approaches were adopted for the control of soft robots, such as soft manipulators \cite{gupta2016learning} and wheeled soft snake robots \cite{liu2023reinforcement}.

As a typical challenge for RL, especially in tasks where complicated behaviors are involved, the learning efficiency suffers from the relatively large search space and the inherent difficulties of the tasks, which usually requires delicate reward shaping \cite{ng1999policy} to guide the policy optimization and to constrain the learning directions or the behavior styles. The RL agents have to successfully reach their goals for efficient learning before getting lost in numerous inefficient failure trials. Multiple strategies were proposed to address the hard exploration challenge with sparse rewards, including improving the exploration techniques for more versatile trajectories from intrinsic motivations \cite{ostrovski2017count, pathak2017curiosity, choshen2018dora, badia2020never}, and exploiting the information acquired from the undesired trails \cite{andrychowicz2017hindsight, fang2018dher, ding2019goal}. 

Compatible with these techniques that might improve learning efficiency, the motivation of BER proposed for off-policy RL is the human ability to solve problems forward (from the beginning to goal) and backward (from the goal to the beginning) simultaneously, which is different from the standard model-free RL algorithms that mostly rely on forward exploration. For example, in proving a complicated mathematical equation, an effective method is to derive the equation from both sides where the information of both the left-hand side (beginning) and the right-hand side (goal) is utilized, to which the reasoning process and the mechanism of BER are similar. 

In this paper, a BER algorithm is introduced that allows the RL agent to explore bidirectionally, which is compatible with arbitrary off-policy RL algorithms. It is applicable for systems with approximate reversibility and with fixed or random goal setups. After an evaluation of BER with a toy task, it is applied to the locomotion and navigation task of a soft snake robot. The developed algorithm is validated on a physics-based dynamic simulator with a computationally efficient serpentine locomotion model based on the system characteristic. Comprehensive experimental results demonstrate the effectiveness of the proposed RL framework with BER in learning the locomotion and navigation skills of the soft snake robot compared with other state-of-the-art benchmarks, indicating the potential of BER in general off-policy RL and robot control applications.

The remainder of the paper is structured as follows. Section \rom{2} introduces the BER algorithm with an evaluation of a toy task. Section \rom{3} details the BER application in locomotion and navigation of a soft snake robot with the performance comparisons with other benchmarks, and Section \rom{4} concludes the paper.

\section{Back-stepping Experience Replay}
\subsection{Background}
\subsubsection{Reinforcement Learning}
\quad\\
\indent
A standard RL formalism is adopted where an agent (e.g. a robot) interacts with an environment and learns a policy according to the perceptions and rewards. 

In each episode, the system starts with an initial state $\bm{s}_0$ with a distribution of $p(\bm{s}_0)$, and the agent observes a current state $\bm{s}_t \in \mathcal{S} \subseteq \mathcal{R}^n$ in the environment at each time step $t$. Then, an action $\bm{a}_t \in \mathcal{A} \subseteq \mathcal{R}^m$ is generated to control the agent based on the current policy $\pi$ and the observations. Afterward, the system evolves to a new state $\bm{s}_{t+1}$ based on the action and transition dynamics $p(\cdot|\bm{s}_t, \bm{a}_t)$, and a reward $r_t=r(\bm{s}_t, \bm{a}_t, \bm{s}_{t+1})$ is collected by the agent for the learning before the termination of the episode. During the training process, the RL agent learns an optimal policy $\pi^*:\mathcal{S} \rightarrow \mathcal{A}$ mapping states to actions that maximize the expected return. The return is defined as the accumulated discounted reward $R_t = \sum_{i=t}^{\infty} \gamma^{i-t} r_{i}$, where $\gamma$ is a discount factor.

The state value function $V^{\pi}(\bm{s}_t)= \mathop{\mathbb{E}}(R_t|\bm{s}_t)$ represents the expected return starting from state $\bm{s}_t$ following the current policy $\pi$, and the action value function $Q^{\pi}(\bm{s}_t, \bm{a}_t)= \mathop{\mathbb{E}}(R_t|\bm{s}_t, \bm{a}_t)$ represents the expected return starting from the state $\bm{s}_t$ with an immediate action $\bm{a}_t$ by following the current policy $\pi$. All optimal policies $\pi^*$ share the same optimal Q-function $Q^*$, according to the Bellman equation \cite{mnih2015human}:
\begin{equation}
    \Scale[0.91]{
    \begin{split}
    Q^*(\bm{s}_t, \bm{a}_t) = \mathbb{E}_{\bm{s}'\sim p(\cdot\bm{s}_t, \bm{a}_t) }\left[r(\bm{s}_t,\bm{a}_t, \bm{s}') 
     + \gamma \max_{\bm{a}' \in \mathcal{A}} Q^*(\bm{s}', \bm{a}')\right]
    \end{split}
    }
\end{equation}
\subsubsection{Deep Q-Networks (DQN) and Deep Deterministic Policy Gradient (DDPG)}
\quad\\
\indent
The Deep Q-Network (DQN) is a model-free, off-policy RL approach suitable for agents operating in discrete action spaces \cite{mnih2015human}. It typically employs a neural network $Q$ to approximate the optimal Q-function $Q^*$, selecting optimal actions: $\bm{a}^* = \arg \max_{\bm{a} \in \mathcal{A}} Q(\bm{s}_t, \bm{a})$. Exploration is often facilitated by the  $\epsilon$-greedy algorithm. To stabilize training, a \textit{replay buffer} stores transition data ($\bm{s}_t, \bm{a}_t, r_t, \bm{s}_{t+1}$) and is used to optimize $Q$ with a loss $\mathcal{L} = \mathbb{E}(Q(\bm{s}_t, \bm{a}_t)-y_t)$, where the target is calculated by using a periodically updated \textit{target network} $Q_{\text{targ}}$: $y_t = r_t + \gamma \max_{\bm{a} \in \mathcal{A}} Q_{\text{targ}}(\bm{s}_{t+1}, \bm{a})$, and using transitions in the \textit{replay buffer}.  

Deep Deterministic Policy Gradient (DDPG) \cite{lillicrap2015continuous_long} is an off-policy RL algorithm that simultaneously learns a Q-function and a policy, tailored specifically for environments with continuous action spaces. DDPG interweaves the learning process of an approximator to $Q^*$, with an approximator to select $\bm{a}^*$, offering a unique adaptation for continuous action scenarios.

\subsection{Algorithm for BER}
The above classical off-policy RL algorithms often face challenges with systems characterized by sparse rewards or challenging tasks with rewards hard to reshape. In such scenarios, RL agents rarely achieve informative standard forward explorations due to a low success rate in reaching goals in complex problems without precise guidance \cite{andrychowicz2017hindsight}. To address these challenges, we propose a novel Back-stepping Experience Replay (BER) algorithm for tasks with different goals (Alg. 1), designed to enhance the learning efficiency of off-policy RL algorithms. This is achieved by incorporating exploration methods in both forward and backward directions. 

The BER algorithm requires at least an approximate reversibility of the system. This means that from a standard transition $(\bm{s}_t, \bm{a}_t, \bm{s}_{t+1})$, a back-stepping transition $(\bm{s}_{t+1}, \widetilde{\bm{a}_t}, \bm{s}_{t})$ can be constructed, which is similar to a real transition $(\bm{s}_{t+1}, \widetilde{\bm{a}_t}, \bm{s}_{b,t})$ in the environment, i.e., $\bm{s}_{b, t} \approx \bm{s}_t$. The action in the back-stepping transition is calculated as $\widetilde{\bm{a}_t} = f(\bm{s}_t,\bm{a}_t,\bm{s}_{t+1})$, where function $f$ is dependent on the environment. The approximate reversibility is evaluated by a small upper bound $K$ for all transitions during back-stepping calculation: 
\begin{equation}
    \lVert \bm{s}_{b,t} - \bm{s}_{t} \rVert \leq K \cdot \lVert \bm{s}_{t+1} - \bm{s}_t \rVert, K<1
\end{equation}

There exists a perfect reversibility when $K=0$ with a probably complex function $f$, while an approximate reversibility might be achieved with a slightly larger $K$ and a simpler and solvable function $f$.

The idea of BER is simple yet effective: instead of solely relying on forward explorations (navy blue solid line in Fig. 1) from initial states to goals, which depend heavily on the randomness of forward trajectories to reach these goals, RL agents also navigate backward from the goals to the initial states in the tasks (sky blue solid line in Fig. 1). The standard transitions are sampled from the standard forward and backward exploration trajectories (solid lines in Fig. 1), where the initial states of themselves are included. Then, the back-stepping transitions are calculated based on the standard transitions to constitute the reversed trajectories (dashed lines in Fig. 1), where the virtual goals are set to be the original initial state in their corresponding standard trajectories, such that the reversed trajectories are guaranteed to reach their virtual goals and contribute to the learning efficiency.

\begin{figure}[!ht]
\vspace{-10pt}
  \centering
\includegraphics[width=0.38\textwidth]{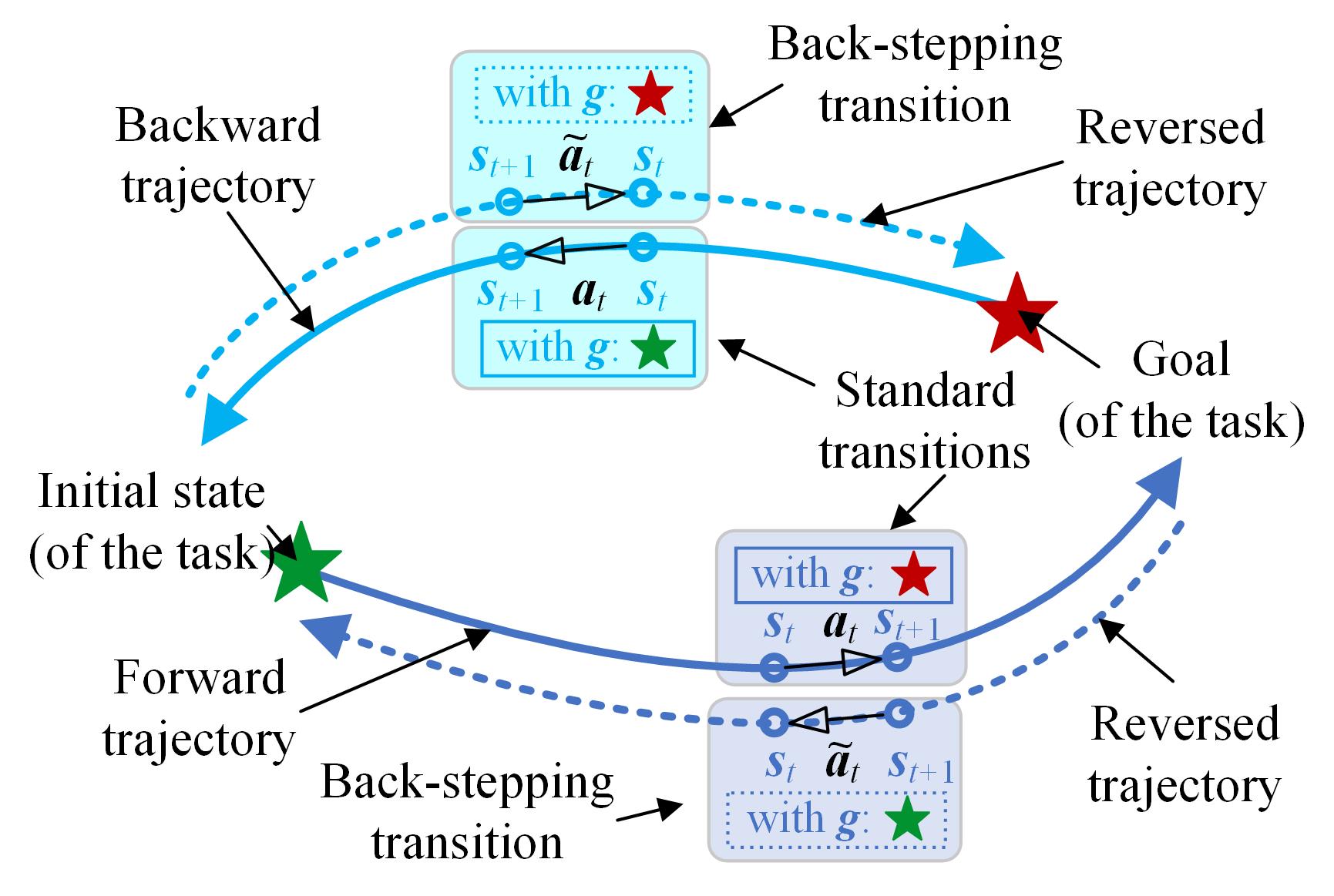}
  \caption{Illustration of the Back-stepping Experience Replay.}
  \label{fig:figure}
  \vspace{-5pt}
\end{figure}

During the explorations, the standard and the back-stepping transitions are collected and stored in separate replay buffers for training. A strategy $\mathbb{S}_t$ is used to sample the transitions from the standard replay $R_f$ with a probability $P_{t,f}$ and from the back-stepping replay $R_b$ with a probability $P_{t,b}$, where $P_{t,f} + P_{t,b} = 1$. For a system with imperfect reversibility, $P_{t,b}$ gradually drops to zero to distill the transition set for training because of the inaccurate back-stepping transition. The details of BER are shown in Alg. 1. It should be noticed that the operator $\odot$ between the states and the goals also indicates the modification of the sequential data (e.g., the history data) when the back-stepping transitions are constructed.

\begin{figure}
\removelatexerror
\scalebox{0.85}{
\begin{algorithm*}[H]
\SetAlFnt{\small}
    \SetKwInOut{Parameter}{Parameter}
    \SetKwInOut{Output}{Output}
    \SetKwInOut{Require}{Require}
    \SetKwInOut{Bullet}{$\bullet$}
\caption{Back-stepping Experience Replay (BER)}
\label{algo:safety_supervisor}
\SetAlgoLined
\nonl\textbf{Given:} \\
\nonl\quad $-$ An off-policy RL algorithm $\mathbb{A}$. \Comment{e.g. DDPG}\\
\nonl\quad $-$ A probability $P_b$ triggering backward trial. \\
\nonl\quad $-$ A strategy $\mathbb{S}_t$ for sampling transitions in replays \\
\nonl\textbf{Require:} \\
\nonl\quad $-$ Approximate reversibility of the system\\
\vspace{0.2em}
\hrule
\vspace{0.2em}
Initialize $\mathbb{A}$  \Comment{e.g. initialize networks}\\
Initialize replay buffers $R_f$ and $R_b$ \\
\For{epoch $=1\rightarrow M$}{
    Sample a goal $\bm{g}$ with an initial state $\bm{s}_0$. \\
    \textbf{Forward trial} starts\\
    \For{t $=0\rightarrow T_{end}-1$}{ 
        Sample an action $\bm{a}_t$ using the policy of $\mathbb{A}$: \\
        $\bm{a}_t \leftarrow \pi(\bm{s}_t \odot \bm{g} )$  \Comment{e.g. $\odot \rightarrow$ diff, concat}\\
        Execute action $\bm{a}_t$, observe new state $\bm{s}_{t+1}$
    }
    \For{$t=0\rightarrow T_{end}-1$}{
        $r_t := r(\bm{s}_t, \bm{a}_t, \bm{s}_{t+1}, \bm{g})$ \\ 
        Store transition $(\bm{s}_t \odot \bm{g}, \bm{a}_t, r_t, \bm{s}_{t+1} \odot \bm{g})$ in $R_f$\\
          \nonl\Comment{standard experience replay} \\
        \nonl\textbf{Construct a back-stepping transition}: \\
        $r_{b,t} := r(\bm{s}_{t+1}, \widetilde{\bm{a}_t}, \bm{s}_{t}, \bm{s}_0)$ \\ 
        Store transition $(\bm{s}_{t+1} \odot \bm{s}_0, \widetilde{\bm{a}_t}, r_{b,t}, \bm{s}_{t} \odot \bm{s}_0)$ in $R_b$
        \nonl\Comment{BER}
        
    }
    \textbf{Forward trial} ends\\
    \textbf{Backward trial} starts with $P_b$ \\
    Swap the goal $\bm{g}$ and the initial state $\bm{s}_0$:
    $\bm{s}_0, \bm{g} = \bm{g}, \bm{s}_0$ \\
    Repeat line 6 - line 16 \\
    \textbf{Backward trial} ends

    \For{$t = 1 \rightarrow N$}{
        Sample a mini-batch $B$ from the replay buffers $\{R_f, R_b\}$ using $\mathbb{S}_r$ \\
        Perform one step of optimization using $\mathbb{A}$ and mini-batch $B$
    }
    
}
\end{algorithm*}}
\vspace{-15pt}
\end{figure}

The BER accelerates the estimation of Q-functions of the RL agent by using the reversed successful trajectories to bootstrap the networks. One interpretation of BER is a bi-directional search method for standard off-policy RL approaches, with a higher convergence rate and learning efficiency. The distillation strategy of the transitions for training needs to be carefully tuned and might be combined with other exploration tricks, to reach an accurate policy learning in the end and avoid the limitations brought by the bi-directional search method, e.g., non-trivial sub-optimum. 

In the practical learning tasks, the accuracy and the complexity of the function $f: \widetilde{\bm{a}_t} = f(\bm{s}_t,\bm{a}_t,\bm{s}_{t+1})$, which calculates the actions $\widetilde{\bm{a}_t}$ in the back-stepping transitions $(\bm{s}_{t+1}, \widetilde{\bm{a}_t}, \bm{s}_{t})$, need to be balanced. An accurate $f$ yields better reversibility (with smaller $K$ in Eq. (2)) with more accurate back-stepping transitions and brings less bias and noise, while $f$ itself could be computationally expensive or even unsolvable. On the other hand, a moderate relaxation of the accuracy of $f$ might boost the efficiency of the calculation of back-stepping transitions, when the larger bias and the noises brought by the approximate reversibility (with larger $K$) are managed by the distillation mechanism in BER.
\subsubsection{A case study of BER}
\quad \\
To illustrate the effectiveness and generality of BER, a general binary bit flipping game \cite{andrychowicz2017hindsight} with $n$ bits was considered as an environment for the RL agent, where the state was the bit value array $\bm{s}  = \{s_i\}_{i=1}^n \in \mathcal{S}$, $s_i \in \{0, 1\}$, and the action was the index of the chosen bit $\bm{a} \in \{1, ..., n \} = \mathcal{A}$ that was flipped. It was noticed that the game was completely reversible and $\widetilde{\bm{a}_t}=f(\bm{a}_t)=\bm{a}_t$ for any time step and transition. The initial state $\bm{s}_0 \in \mathcal{S}$ and the goal $\bm{g}\in \mathcal{S}$ were sampled uniformly and randomly, with a sparse non-negative reward: $r_t(\bm{s}, \bm{a}) = -[\bm{s} \neq \bm{g}]$. The game is terminated once $\bm{s} = \bm{g}$.

A simple ablation study was designed where a DQN and a DQN with BER were used for training when $n = 4, 6, 8$. The fully activated backward exploration and the use of back-stepping transitions were stopped after 1k epochs directly. The experimental result (Fig. 2) showed that BER facilitated an effective and efficient policy learning for a general DQN approach, and contributed more when the problem became more complex (i.e., $n$ was larger). 

\begin{figure}[!ht]
\vspace{-10pt}
  \centering
\includegraphics[width=0.495\textwidth]{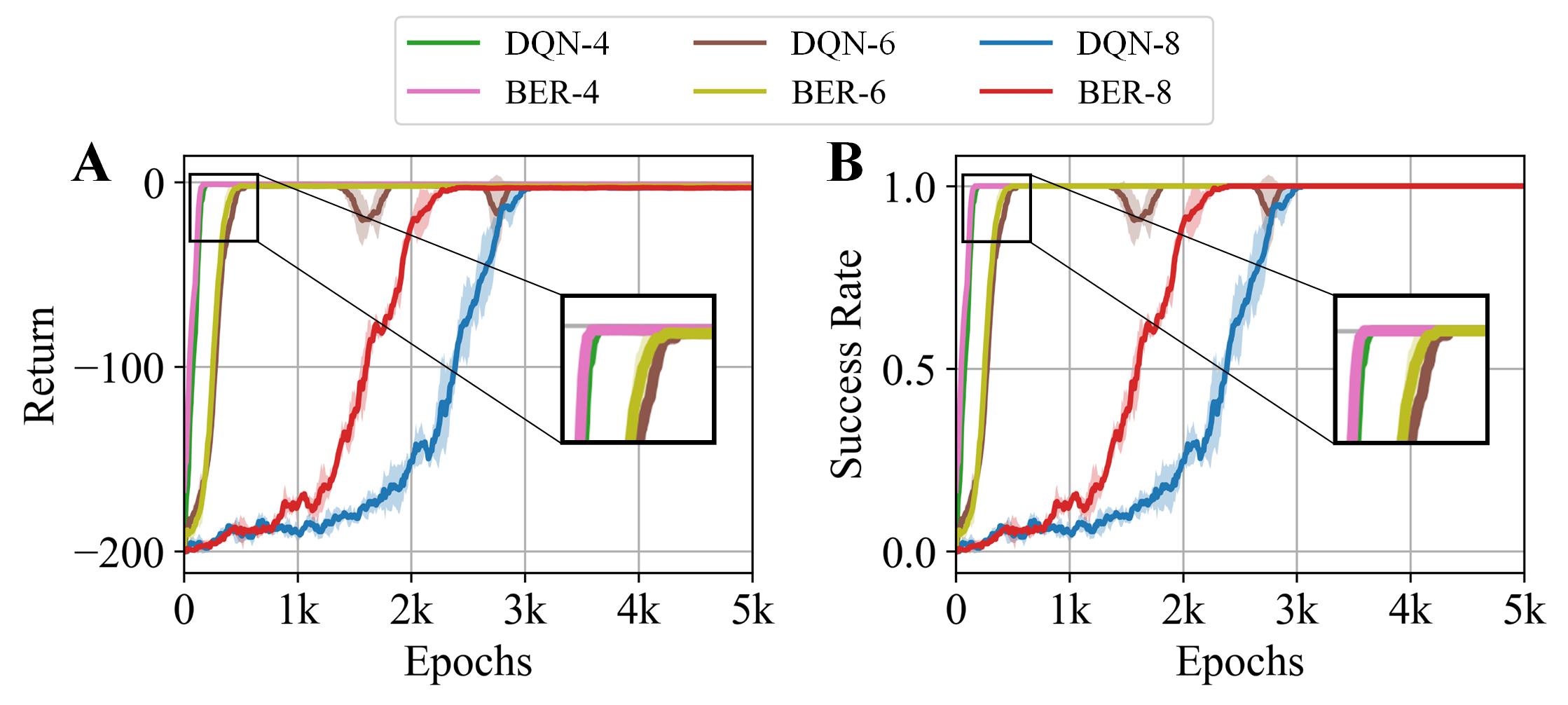}
  \vspace{-15pt}
  \caption{Training experiments of the bit flip game with different algorithms and state dimensions. (\textbf{A}) Returns; (\textbf{B}) Success rates.}
  \label{fig:figure}
  \vspace{-10pt}
\end{figure}

\section{BER in Model-free RL for a Soft Snake Robot}
In this section, a locomotion and navigation task for a compact pneumatic soft snake robot with snake skins in our previous works \cite{qi2020novel, qi2023bioinspired} is utilized to evaluate the effectiveness and efficiency of BER with a model-free RL approach, where the robot learns both movement skills and efficient strategies to reach different challenging targets. 


\subsection{Soft Snake Robot and Serpentine Locomotion}
Compared with soft snake robots where each air chamber was controlled independently \cite{luo2014theoretical}, in this paper, a more compact soft snake robot with snake skins \cite{qi2020novel} is considered. There are only four independent air paths to generate the traveling-wave deformation of the robot, which enables the robot to traverse complex environments more easily by reducing the number of pneumatic tubing. The body of the robot consists of six bending actuators and each actuator is divided into four air chambers (Figs. 3A, 3D) that connect to four air paths (Fig. 3B). Four sinusoidal waves with 90-degree phase differences and the same amplitude can be used as references of pressures in air paths to generate traveling-wave deformation (Fig. 3C), when the biases of waves induce unbalance actuation for steering of the robot.

\begin{figure*}[!ht]
  \centering
\includegraphics[width=0.9\textwidth]{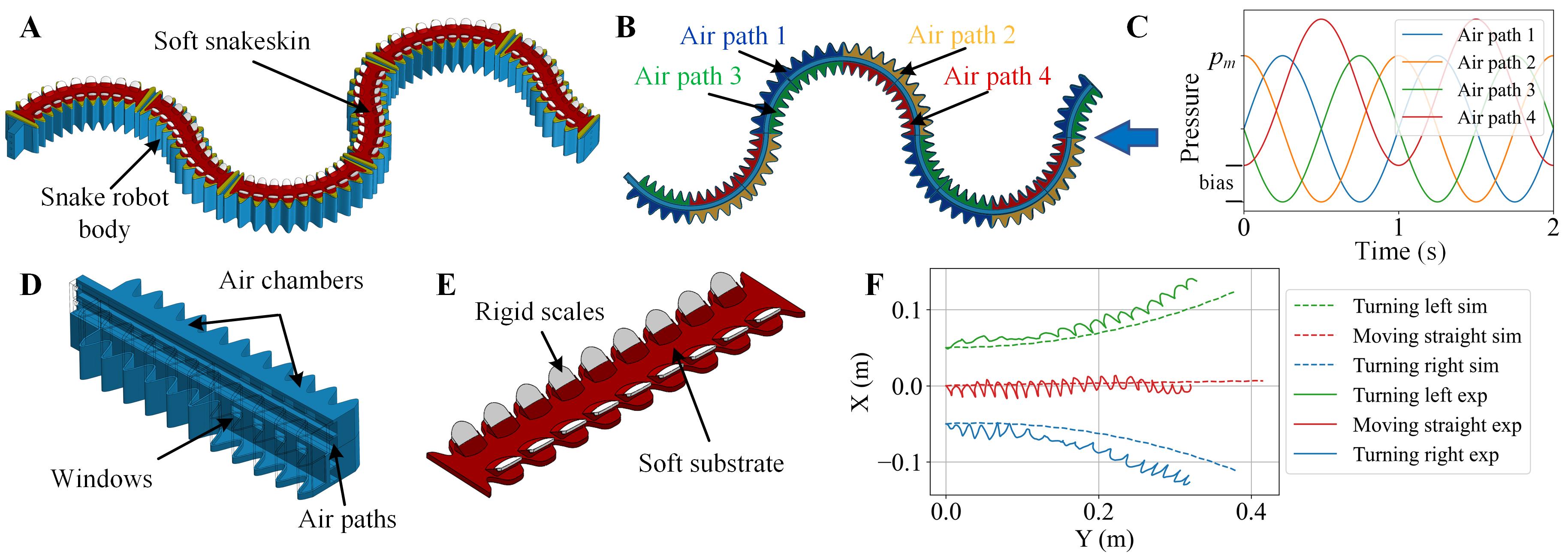}
  \vspace{-3pt}
  \caption{The overview of the soft snake robot with skins. (\textbf{A}) The soft snake robot with soft snakeskins; (\textbf{B}) The connection between air chambers and air paths; (\textbf{C}) The actuation pressures for air paths; (\textbf{D}) The structure of one bending actuator; (\textbf{E}) The structure of soft snakeskin; (\textbf{F}) The simulation (sim) and experimental (exp) results of the trajectory of the COM of the snake robot on a rough paper surface.}
  \label{fig:figure}
  \vspace{-12pt}
\end{figure*}

Serpentine locomotion is adopted for the movement of the soft snake robot, where the anisotropic friction between the snake skins and the ground propels the robot during the traveling-wave deformation \cite{hu2009mechanics}. The artificial snake skins are designed with a soft substrate and embedded rigid scales (Fig. 3E); see \cite{qi2023bioinspired} for more details.

To describe the serpentine locomotion of the robot, the dynamic model in \cite{hu2009mechanics} is adopted, where the body of the robot is modeled as an inextensible curve in a 2D plane with a total length $L$ and a constant density $\rho$ per unit length. The position of each point on the robot at time $t$ is defined as:
\begin{equation}
    \bm{X}(s, t) = (x(s, t), y(s, t))
\end{equation}
where $s$ is the curve length measured from the tail of the robot.

By utilizing a mean-zero anti-derivative $I_0$ \cite{hu2012slithering} ($I_0[f](s,t)$ $= \int_0^s f(s',t)\dd{s}' -\frac{1}{L}\int_0^L\dd{s}\int_0^s\dd{s'}f(s',t) $), the position $\bm{X}(s,t)$ and the orientation $\theta(s,t)$ (the angle between the local tangent direction and the \textit{X}-axis of the inertial frame) of each point are described as a function of the position $\overline{\bm{X}}(t)$ and orientation $\overline{\theta}(t)$ (Fig. 4) of the center of mass (COM) of the robot:
\begin{equation}
    \bm{X}(s, t) = \overline{\bm{X}}(t) + I_0[\bm{X}_s](s,t)
\end{equation}
\begin{equation}
    \theta(s, t) = \overline{\theta}(t) + I_0[\kappa](s,t)
\end{equation}
where $\bm{X}_s = (\cos{\theta}, \sin{\theta})$ and $\kappa(s,t)$ is the local curvature. $\overline{\bm{X}}(t) = \frac{1}{L} \int_0^L \bm{X}(s, t) \dd{s}$, $\overline{\theta}(t) = \frac{1}{L} \int_0^L \theta(s, t) \dd{s}$. The curvature $\kappa(s,t)$ is related to the local pneumatic pressure via:
\begin{equation}
    \kappa(s,t) = K_b \cdot \Delta p(s,t)
\end{equation}
where $K_b$ is the proportional constant and $\Delta p(s,t)$ is the pressure difference between the two air chambers at point $s$.

\begin{figure}[!ht]
\vspace{-5pt}
  \centering
\includegraphics[width=0.32\textwidth]{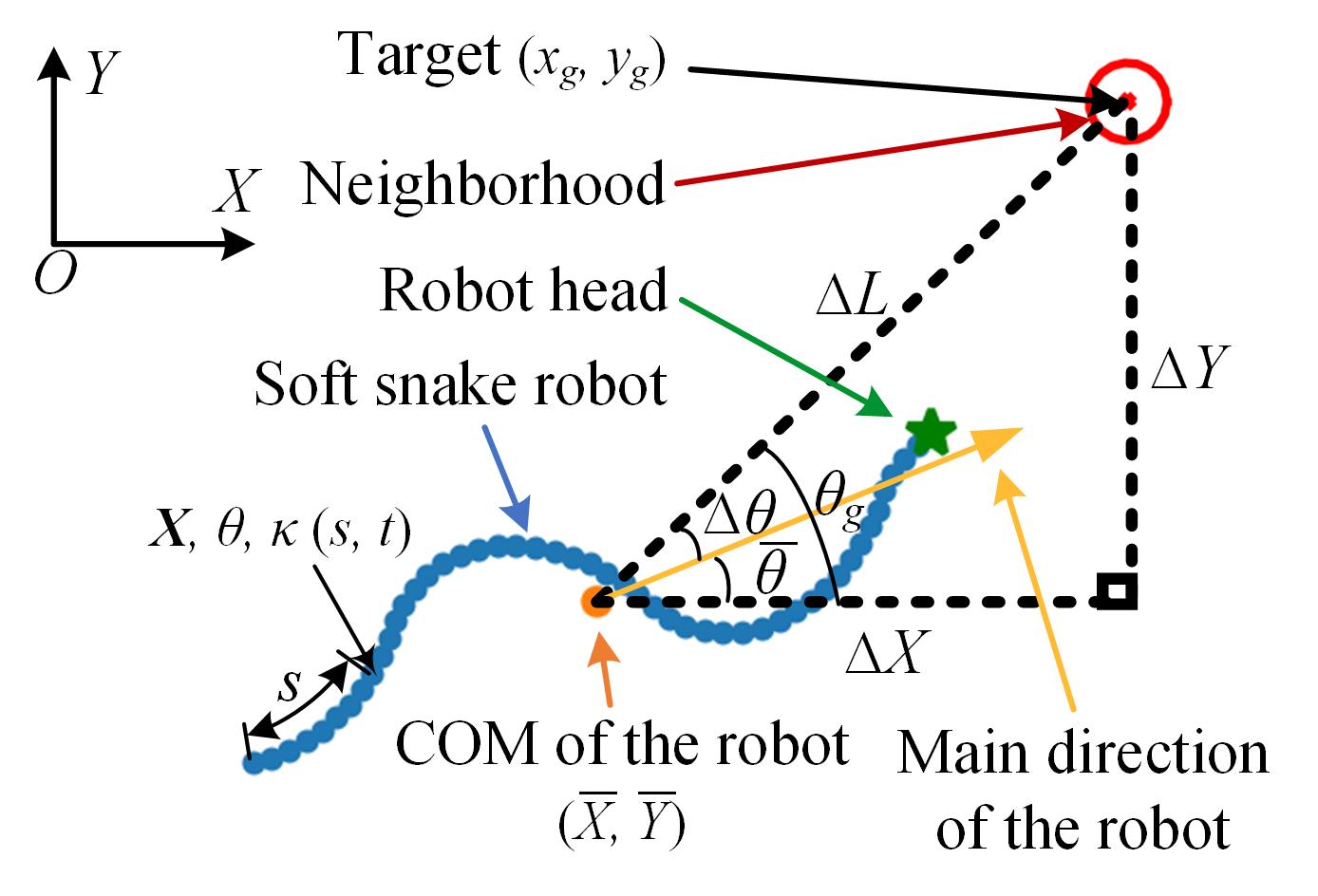}
  \vspace{-3pt}
  \caption{The illustration of the soft snake robot with serpentine locomotion approaching a target.}
  \label{fig:figure}
  \vspace{-10pt}
\end{figure}

The anisotropic friction $\bm{f}_{fric}$ between the snake skins and the ground is described as a weighted average of the independent components in different local directions (forward $\hat{\bm{f}}$, backward $\hat{\bm{b}}$, transverse $\hat{\bm{t}}$):
\begin{equation}
\begin{cases}
    \bm{f}_{fric} = -\rho g (\mu_t(\hat{\bm{u}}\cdot\hat{\bm{t}})\hat{\bm{t}} + \mu_l(\hat{\bm{u}} \cdot \hat{\bm{f}}) \hat{\bm{f}} ) \\
    \mu_l = \mu_f H(\hat{\bm{u}}\cdot\hat{\bm{f}}) + \mu_b(1-H(\hat{\bm{u}}\cdot\hat{\bm{f}}))
\end{cases}
\end{equation}
where $\hat{\bm{u}}$ represents the direction of the local velocity, $\mu_f$, $\mu_b$, and $\mu_t$ are the friction coefficients of the snakeskin in $\hat{\bm{f}}$, backward $\hat{\bm{b}}$, and $\hat{\bm{t}}$ directions, respectively. $H(x)=(1+sgn(x))/2$, where $sgn$ is the signum function.

The dynamics of each point of the snake robot is determined by Newton’s second law:
\begin{equation}
    \rho \Ddot{\bm{X}} = \bm{f}_{fric} + \bm{f}_{inte}
\end{equation}
where $f_{inte}$ is the internal force in the robot body, which includes internal air pressure, bending elastic force, et al, with observations: $\int_0^L \bm{f}_{inte} = 0$ and $\int_0^L (\bm{X}(s,t) - \overline{\bm{X}}(t)) \times \bm{f}_{inte} = 0$.

Finally, the dynamics for the COM of the robot are derived using the equation (3)-(8) with the observations of $f_{inte}$; see \cite{hu2012slithering} for more details.

Based on the dynamic model of the robot, which simplifies a dynamic system for all points of the robot to a single dynamic system for the COM of the robot, a simulator is designed with proper discretizations and numerical techniques for RL training. The simulation results matched the experimental results \cite{qi2023bioinspired} of the soft snake robot when different pressure biases were applied for the robot's steering (Fig. 3F), where the wavy trajectories in the experiments were attributed to the limited number (25) of the tracking markers in the tests. 

\subsection{RL formulation of Locomotion and Navigation of the Robot}
In this paper, the locomotion and navigation of the soft snake robot is formulated as a Markov Decision Process (MDP) $\mathcal{M}$ and solved with a model-free RL. The $\mathcal{M}$ is defined as a tuple $\mathcal{M} = (\mathcal{A}, \mathcal{S}, \mathcal{R}, \mathcal{T}, \gamma)$:
\begin{enumerate}[wide, labelwidth=!, labelindent=10pt]    
     \item \textbf{Action space}: Compared with a random Central Pattern Generator (CPG) \cite{liu2023reinforcement}, more constrained sinusoidal waves are used to generate a smoother traveling-wave deformation of the robot for better locomotion efficiency. Besides, the learned controller of the robot is limited to avoid high-frequency pressure change, i.e., the RL agent is only able to generate an action to change the parameters of the waveform at the beginning of each actuation period $[0, T]$ that is same as the period of the sinusoidal waves, and one episode consists of multiple connected actuation periods. The sinusoidal pressure $p_i$ for $i$-th channel of the robot is designed as:
    \begin{equation}
    \Scale[1]{
    \begin{split}
        p_i = p_m\sin{( c \cdot \frac{2\pi}{T} t_r + \frac{(i-1)\cdot\pi}{2})} + b_{i, pre} + (b_i - b_{i, pre}) \frac{t_r}{T}
    \end{split}
    }
    \end{equation}
    where $t_r \in [0,T]$ is the relative time in one actuation period. $p_m$ and $b_i \in [0, b_m]$ are the fixed magnitude and bias of the sinusoidal waves for the $i$-th channel, respectively, $i \in \{1,2,3,4\}$. $b_{i, pre}$ is a one-step history of the wave bias $b_i$ for the $i$-th channel with $b_{i, pre} = 0$ at the initial state. $c \in \{-1, 1\}$ is a variable to control the propagation direction of the traveling-wave deformation and thus can change the movement direction of the robot.
    
    The action space $\mathcal{A}$ of the RL agent for locomotion and navigation of the robot is designed as:
    \begin{equation}
        \bm{a} = \{b_{a,1}, b_{a,2}, c\} \in \mathcal{A}
    \end{equation}
    where $b_i$'s are constructed by $b_{a,1} \in [-b_m, b_m]$ and $b_{a,2} \in [-b_m, b_m]$: 
    \begin{equation}
        \begin{cases}
            b_1, b_3 = \max(0, b_{a,1}),  -\min(0, b_{a,1}) \\
            b_2, b_4 = \max(0, b_{a,2}),  -\min(0, b_{a,2})
        \end{cases}
    \end{equation}

    At the beginning of each actuation period, based on the current policy, the RL agent observes the state and generates an action, which specifies the waveform of the pressures in that period to propel the snake robot. The wave design guarantees the continuity of the pressures across different actuation periods to avoid impractical sudden changes in the pressures and the robot's body shape.

    \item \textbf{State space}: A goal-conditioned state is used for the learning of the RL agent for adapting to different random targets. Specifically, a relative representation of the snake robot's position and orientation with respect to the target is used as part of the state (Fig. 4):
    \begin{equation}
         \bm{s} = \{\Delta X, \Delta Y, \Delta \theta, \bm{b}_{a,1,pre}, \bm{b}_{a,2,pre}\} \in \mathcal{S}
    \end{equation}
    where $\Delta X = x_g - \overline{X}$, $\Delta Y = y_g - \overline{Y}$ denote the relative position of the target to the COM of the snake robot, $\Delta \theta = \theta_g - \overline{\theta} \in (-\pi, \pi]$ represents the relative direction of the target to the main direction of the robot, and $\theta_g = \arctan(\Delta Y / \Delta X)$ is the angle between the line from the COM of the robot to the target and the \textit{X}-axis, $\bm{b}_{a,1, pre}$ and $\bm{b}_{a,2, pre}$ are two-step histories of the action $b_{a,1}$ and $b_{a,2}$, respectively, with an initial setting of $\{0,0\}$.

    The velocities of COM of the robot are not included as part of the state because the value of the Froude number $Fr$ \cite{hu2012slithering} in serpentine locomotion of the snake robot is small, indicating that the frictional and gravitational effects dominate the inertial effect. Two-step histories (longer than one step) are introduced to compensate for the omission of the velocity state.

    


    
    \item \textbf{Reward function}: The reward function $r$ is pivotal for the RL agent to learn the desired behaviors. The training objective in this work is to drive the COM of the snake robot to reach a random target as soon as possible, with a preference for serpentine locomotion where the robot approaches the target along its main direction. Therefore, the reward assigned to the agent at time $t$ is designed as:
    \begin{equation} \label{eqn:reward_fn}
    r_{t} = 
    \begin{cases}
    \begin{split}
        &w_1 \frac{\Delta L_t}{\Delta L_0} +w_2 \frac{2 \Delta \theta_{r, t}}{\pi} + R_g, & \Delta L_t \leq \epsilon \\
        &w_1 \frac{\Delta L_t}{\Delta L_0} +w_2 \frac{2 \Delta \theta_{r, t}}{\pi}, & \text{else}
    \end{split}
    \end{cases}
    \end{equation}
    where $w_1$ and $w_2$ are non-positive coefficients, $R_g$ is a large sparse positive success reward once the COM of the robot enters a neighborhood of the target with a radius of $\epsilon$. $\Delta L_t = \sqrt{\Delta X_t^2 + \Delta Y_t^2}$ is the distance between the COM of the robot and target at time $t$, and $\Delta L_t = \Delta L_0$ when $t=0$. The deflection $\Delta \theta_{r, t} \in [0, \pi/2]$ is used in the reward to allow the robot to approach the target in a backward direction as well:
    \begin{equation}
        \Delta \theta_{r, t} =
        \begin{cases}
        \begin{split}
            &|\Delta \theta_t|, & -\pi/2 \leq \Delta \theta_t \leq \pi/2 \\
            &\pi - |\Delta \theta_t|, & \text{else}
        \end{split}
        \end{cases}
    \end{equation}
    
    \item \textbf{Transition probabilities}: The transition probability, $\mathcal{T}(\bm{s}'|\bm{s}, a)$, characterizes the underlying dynamics of the robot system in the environment. In this study, we do not assume any detailed knowledge of this transition probability while developing our RL algorithm.
\end{enumerate}

\subsection{Experiments of RL algorithms}
\subsubsection{Experimental Setups}
\quad \\
The RL experiments for the locomotion and navigation of the snake robot were conducted in a customized dynamic simulator which was developed based on the aforementioned serpentine locomotion model (Sec. III. A). The soft snake robot had a length of 0.5 m with a linear density of 1.08 kg/m. The frictional anisotropy between the snake skins and the ground was set as $\mu_f: \mu_b: \mu_t = 1:1:1.5$, and the maximum of the pressure bias $b_m$ was set as the same as $p_m = 276$ kPa. The proportional constant $K_b$ between the applied pressure difference and the curvature was set as 0.058 kPa$\cdot$m. The period of the actuation and the sinusoidal waves was 1 s.

The serpentine locomotion of the soft snake robot demonstrated approximate reversibility (Fig. 5A) when the function $f$ was designed as: $\widetilde{\bm{a}_t} = f(\bm{a}_t) = \{b_{a,1}, b_{a,2}, -c\}$ when $\bm{a}_t = \{b_{a,1}, b_{a,2}, c\}$. The trajectories in the simulation results (Fig. 5A) suggested a small $K<1$ (in Eq. (2)) for locomotion and navigation of the soft snake robot when the above function $f$ was used.

\begin{figure}[!ht]
\vspace{-5pt}
  \centering
\includegraphics[width=0.49\textwidth]{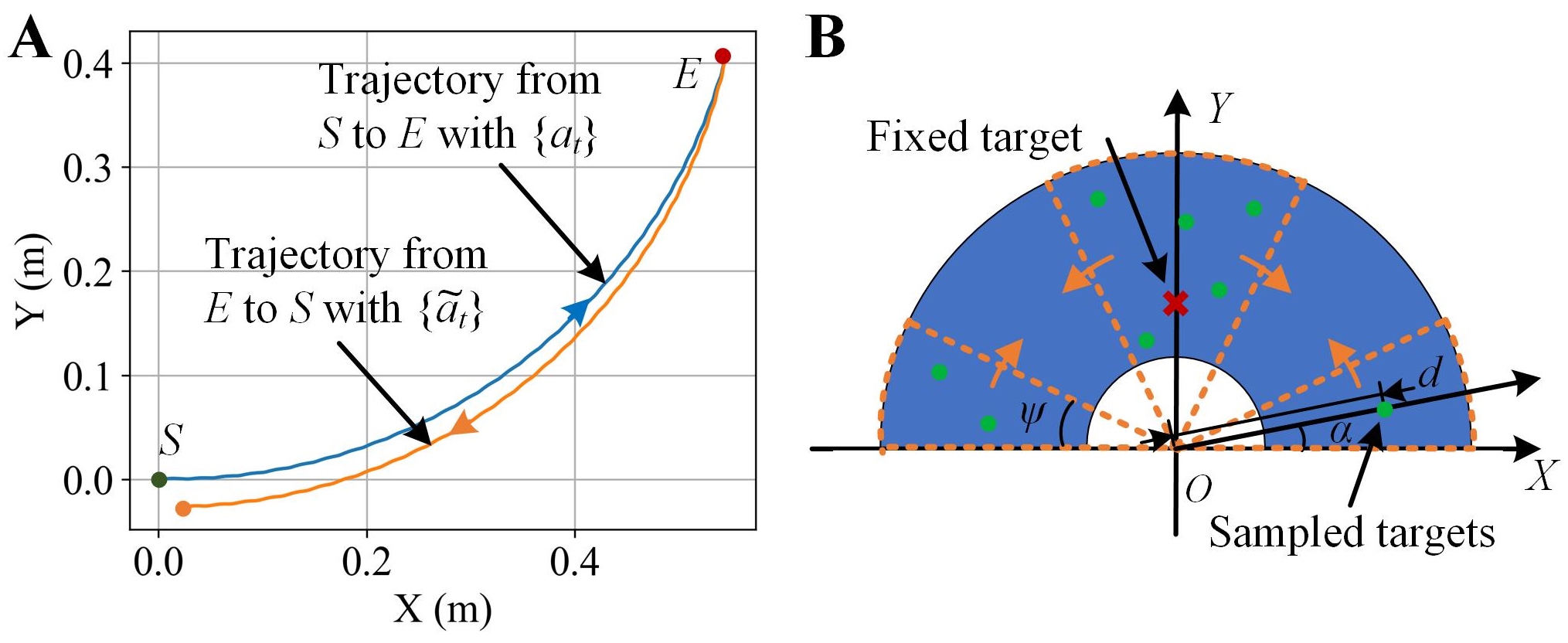}
  \caption{(\textbf{A}). The approximate reversibility of the movement of the soft snake robot with snake skins. (\textbf{B}). The fixed target and the sampling range of the random targets.}
  \label{fig:figure}
  \vspace{-5pt}
\end{figure}

\begin{figure*}[h!]
  \centering
\includegraphics[width=0.98\textwidth]{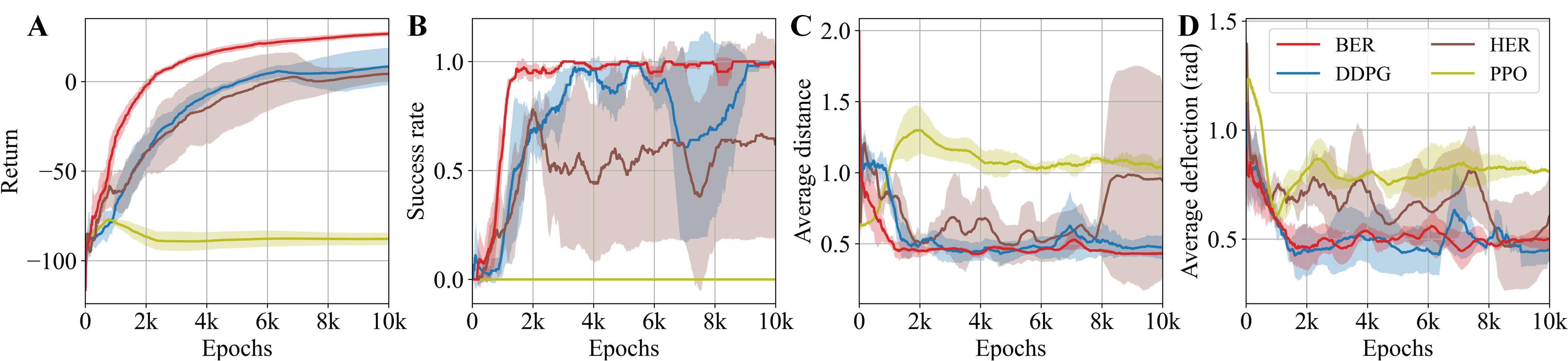}
  \caption{Experimental results of the training for locomotion and navigation of the soft snake robot with one fixed target $(0, 0.5$ m$)$. (\textbf{A}) Returns; (\textbf{B}) Success rates; (\textbf{C}) Average distances; (\textbf{D}) Average deflections.}
  \label{fig:figure}
  \vspace{-10pt}
\end{figure*}

\begin{figure*}[h!]
  \centering
\includegraphics[width=0.98\textwidth]{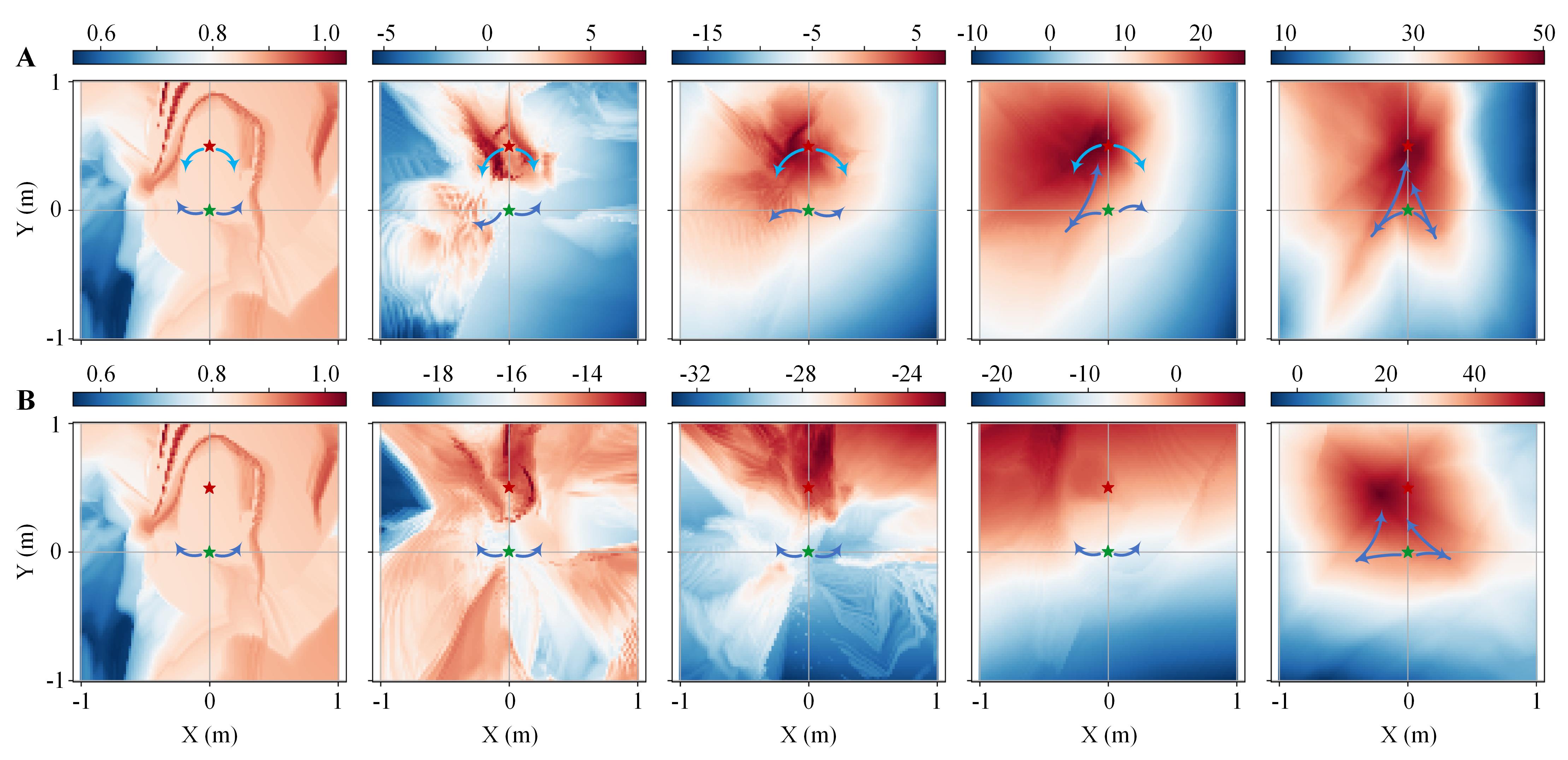}
\vspace{-5pt}
  \caption{Evolution of the maximum Q-value at different locations during the training (from left to right: initial state, epoch 100, 500, 1k, 10k). (\textbf{A}) Training with BER; (\textbf{B}) Training with DDPG. }
  \label{fig:figure}
  \vspace{-12pt}
\end{figure*}

The soft snake robot was initialized in the simulator by using a horizontal static curved shape  ($(\overline{X}, \overline{Y})=(0,0), \overline{\theta}=0$) with zero-value action histories and a target (with neighborhoods: $\epsilon = 0.03$ m), whose control policies was learned by using BER (with DDPG) and several state-of-art benchmark algorithms, including DDPG, HER \cite{andrychowicz2017hindsight}, and PPO \cite{schulman2017proximal}. The number of total training epochs was 10k and the strategy to sample the transitions was $P_{t,b} = 0.5 e^{-0.002i}$ when the index of epoch $i \leq 2500$, $P_{t,b} = 0$ when $i>2500$, and $P_{t,f} = 1-P_{t,b}$, $P_b = P_{t,b}$. The coefficients of the reward were set as $\omega_1 = 0.15$, $\omega_2 = 1$, and the termination condition for one episode was either the COM of the robot entering a neighborhood of the target and receiving a success reward ($R_g = 50$)  or the exploration time exceeding 150 s.

The return, success rate, average distance (the averaged $\Delta L_t / L_0$ for each time step $t$), and average deflection (the averaged $\Delta \theta_{r, t}$ for each time step $t$) were used to evaluate the algorithms during the training, with moving-window averaging for training with different seeds ($l_{window} = 50$ epochs). Three training experiments with different random seeds (for parameter initialization) were conducted to evaluate each algorithm, where the solid line and the shaded area showed the mean and the standard deviation, respectively (Figs. 6 and 8). An AMD 9820X processor with 64 GB memory and Ubuntu 18.04 was used for the training.

\subsubsection{Locomotion and Navigation with a Fixed Target}
\quad \\
The performance of the algorithms was initially evaluated on the locomotion and navigation task of the robot, targeting a challenging fixed point $(x_g, y_g) = (0, 0.5~m)$ (Fig. 5B). The experiment results of the training showed that both DDPG and BER were able to solve the task and learn policies to reach the fixed target successfully, while HER had worse stability and PPO was unable to solve the task within the epoch limitation (Fig. 6). It was also shown that BER had a faster convergence rate and better stability compared with other baseline algorithms.
\begin{figure*}[!ht]
  \centering
\includegraphics[width=0.98\textwidth]{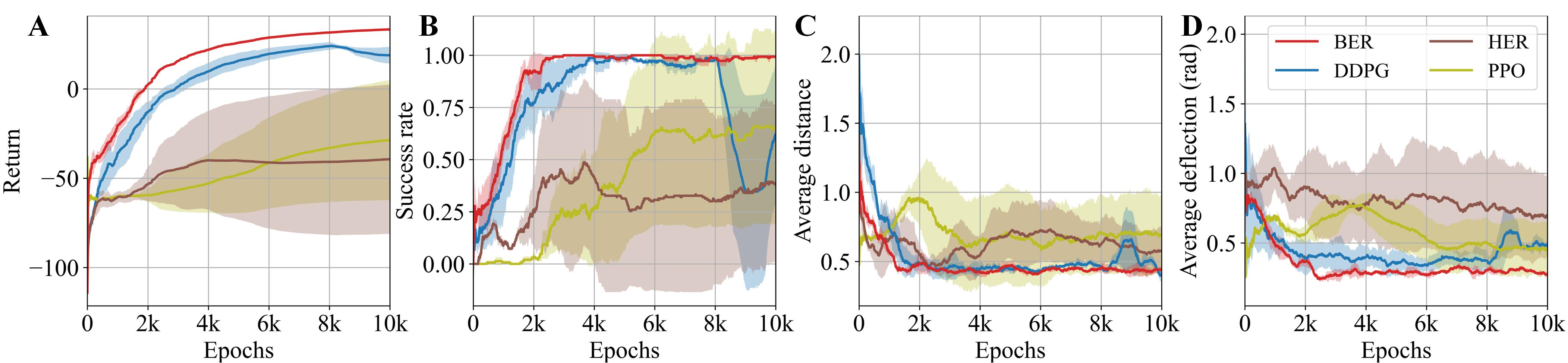}
  \caption{Experimental results of the training for locomotion and navigation of the soft snake robot with random targets. (\textbf{A}) Returns; (\textbf{B}) Success rates; (\textbf{C}) Average distances; (\textbf{D}) Average deflections.}
  \label{fig:figure}
  \vspace{-10pt}
\end{figure*}

The evolution of the maximum Q-value at different locations for the algorithms during the training process (with the same seed) revealed the underlying mechanism and the advantage of BER (Fig. 7). It was shown that the effective Q-values in the training with BER were estimated from both the start and the target locations, expediting the successful explorations and the convergence of the estimation. The BER learned a more informative Q-value distribution after 500 epochs than that of the baseline DDPG after 1000 epochs. The final Q-value distribution of BER was also more accurate than that of the baseline DDPG, manifested by their shapes and the positions of the Q-value's peaks.

\subsubsection{Locomotion and Navigation with Random Targets}
\quad \\
A locomotion and navigation task of the soft snake robot with random targets was then explored by using different RL algorithms, where a half ring was used to randomly sample the target because of the system symmetry: $g \in \{(d, \alpha) \:|\:  d \in [0.3, 1], \alpha \in [0, \pi] \}$ (Fig. 5B). Besides, a strategy was designed where the targets were sampled uniformly from gradually expanding areas for the $i$-th training epoch within the total $n$ epochs: $g \in \{(d, \alpha) \:|\: d \in [0.3, 1], \alpha \in [0, \psi] \cup (\frac{\pi}{2} - \psi, \frac{\pi}{2} +  \psi] \cup (\pi - \psi, \pi] \}$, $ \psi = \frac{\pi}{4n^2} i^2$.

The training results revealed that BER outperformed all other benchmark algorithms tested (Fig. 8). BER achieved the highest return and success rate during training, exhibiting more stable behavior and a smaller average deflection. In contrast, the baseline DDPG's performance declined when introduced to a variety of targets, despite its strong early-stage performance. HER struggled to learn to reach targets in different areas, whereas PPO gradually learned an effective policy, a process that benefited from the random-goal training setup involving progressively changing targets.

The robot's trajectories further demonstrated BER's efficiency (Fig. 9), where controllers with median success rates from each algorithm were used for control. A video for these experiments can be viewed at \url{https://youtu.be/Z0da6rVu9j8}. Three representative targets were tested: $(-0.8$ m$, 0.1$ m$)$ for moving backward, $(0, 0.5$ m$)$ for moving towards a lateral target, $(0.8$ m$, 0.1$ m$)$ for moving forward. The BER controller successfully and smoothly guided the robot to all targets. In contrast, the DDPG and HER controllers exhibited inefficient oscillations, possibly due to less accurate Q-function estimation. While the PPO controller managed to reach all targets, it also displayed inefficient oscillation and adopted a sub-optimal policy for the forward target $(0.8$ m$, 0.1$ m$)$.

\begin{figure*}[h!]
  \centering
\includegraphics[width=0.95\textwidth]{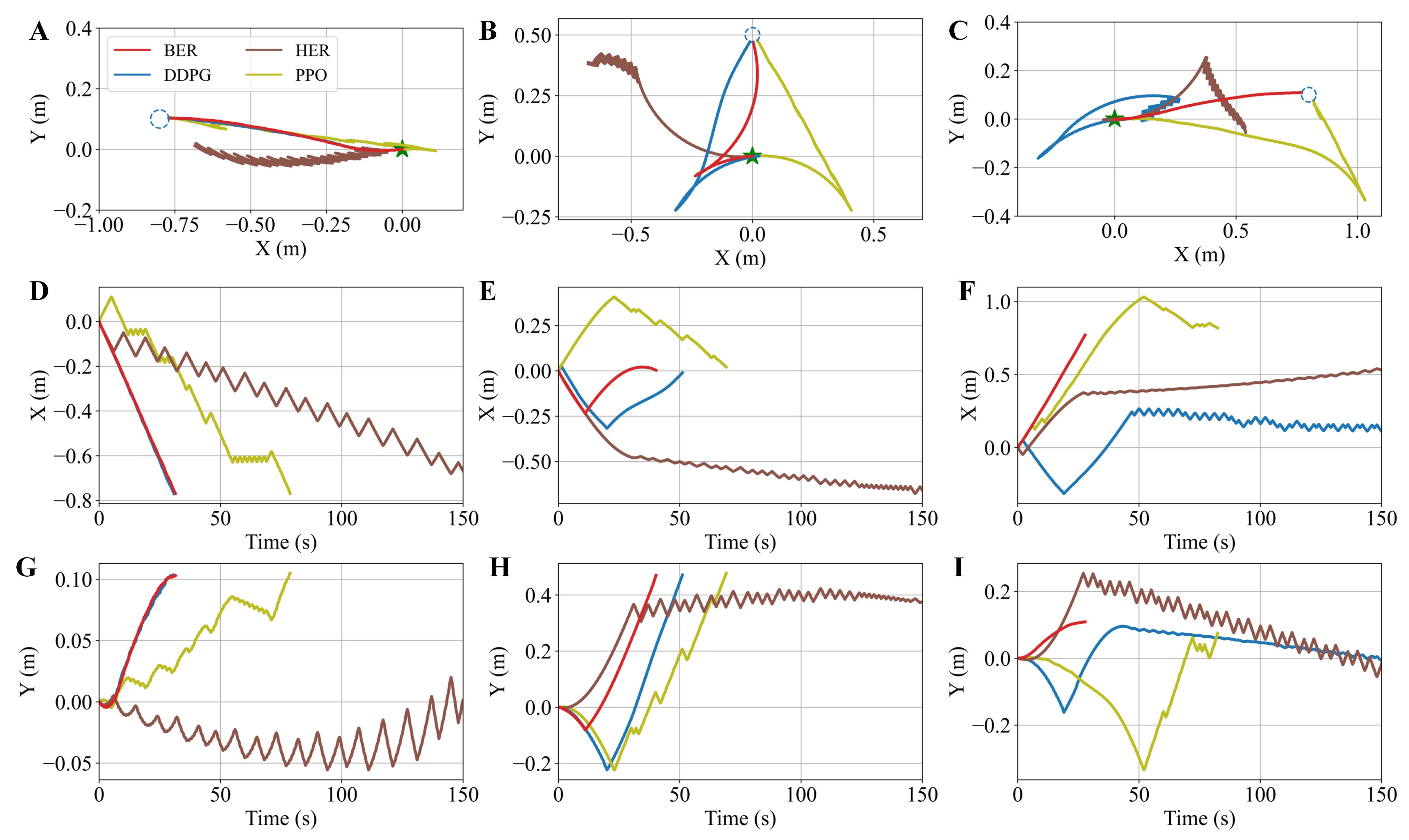}
  \caption{Trajectories of the COM of the soft snake robot by using the controllers learned by different algorithms. (\textbf{A}) Trajectories with a backward target where the relationships between positions and time are shown in (\textbf{D}), (\textbf{G}); (\textbf{B}) Trajectories with a lateral target where the relationships between positions and time are shown in (\textbf{E}), (\textbf{H}); (\textbf{C}) Trajectories with a forward target where the relationships between positions and time are shown in (\textbf{F}), (\textbf{I}). 
  }
  \label{fig:figure}
  \vspace{-10pt}
\end{figure*}
The quantitative results of the algorithms (Table I) were the average values tested by using the controllers trained with different seeds, and using 50 random targets sampled from the half-ring area (Fig. 5B). The average velocity ($v_{avg} = \Delta L_{0} / t_{ep}$, $t_{ep}$: episode length) indicated the efficiency of the learned controllers. Notably, the average velocity of the robot with the BER controller (0.0169 m/s) was approximately 48\% faster than that of the DDPG baseline (0.0114 m/s), and significantly higher compared to other benchmarks.

Besides, compared to other algorithms, BER not only learned an efficient controller based on the primary reward (highest average deflection: $0.2920$ rad), but was also able to sacrifice the secondary reward to some extent (second highest average distance: $0.4002$ m/m) for better performance. The success rate of BER reached 100$\%$ while that of the other baselines did not exceed 65$\%$, which exhibited the advantage of BER in the locomotion and navigation learning of the soft snake robot.

\begin{table}[!ht]
\renewcommand{\arraystretch}{1.4}
\centering
\caption{Testing performance comparisons of different algorithms.}
\label{tab:wc}
\begin{tabular}{lcccc}
\hline 
\multicolumn{1}{c}{\textbf{Metrics}}              & \textbf{PPO} & \textbf{HER} & \textbf{DDPG}  & \textbf{BER} \\ \hline \hline 
Average velocity (m/s) & 0.0061  &  0.0080 & 0.0114  & \textbf{0.0169} \\ 
Average distance (m/m) & 0.6241  & 0.5202 & \textbf{0.3903}  & 0.4002 \\ 
Average deflection (rad) &  0.4049  &  0.6702 &  0.3915  &  \textbf{0.2920} \\ 
Success rate (\%)     &  64.44   & 43.33 & 61.11  &  \textbf{100}  \\ \hline 
\end{tabular}
\vspace{-10pt}
\end{table}

\section{Conclusions and Discussions}\label{sec:5}
A novel technique, Back-stepping Experience Replay, was proposed in this paper, which exploited the back-stepping transitions constructed by using the standard transitions in both forward and backward exploration trajectories, improving the learning efficiencies in off-policy RL algorithms for the approximate reversible systems. The BER was compatible with arbitrary off-policy RL algorithms, demonstrated by combining with DQN and DDPG in a bit-flip task and locomotion and navigation task for a soft snake robot, respectively.

A model-free RL framework was proposed for locomotion and navigation of a soft snake robot as an application of the proposed BER, where a conventional locomotion model for real snakes was adopted to describe the serpentine locomotion of the soft snake robot and to design a simulator for learning. An RL formulation for locomotion and navigation of the soft snake robot was built based on the characteristics of the robot. Extensive experiments showed that the proposed RL approach was able to learn an efficient controller that drove the soft snake robot approaching fixed or even random targets by using serpentine locomotion. For the tasks with random targets, the controller learned by using BER achieved a 100 $\%$ success rate and the robot's average speed was 48 $\%$ faster than that of the best baseline RL benchmark.

For future work, we will apply the proposed RL approach with BER to a physical soft snake robot system, where the data from the physical system will be used for learning. Besides, we will also study the influence of the approximate reversibility of general systems to BER, and analyze the convergence properties of BER for proper state-of-the-art off-policy RL algorithms.

\bibliographystyle{ieeetr}
\bibliography{ref.bib}

\begin{thebibliography}{10}

\bibitem{silver2016mastering_long}
D.~Silver, A.~Huang, C.~J. Maddison, A.~Guez, L.~Sifre, G.~Van Den~Driessche, J.~Schrittwieser, I.~Antonoglou, V.~Panneershelvam, M.~Lanctot, {\em et~al.}, ``Mastering the game of go with deep neural networks and tree search,'' {\em Nature}, vol.~529, no.~7587, pp.~484--489, 2016.

\bibitem{liu2020learning}
X.~Liu, R.~Gasoto, Z.~Jiang, C.~Onal, and J.~Fu, ``Learning to locomote with artificial neural-network and cpg-based control in a soft snake robot,'' in {\em 2020 IEEE/RSJ International Conference on Intelligent Robots and Systems (IROS)}, pp.~7758--7765, IEEE, 2020.

\bibitem{chen2020autonomous}
D.~Chen, L.~Jiang, Y.~Wang, and Z.~Li, ``Autonomous driving using safe reinforcement learning by incorporating a regret-based human lane-changing decision model,'' in {\em 2020 American Control Conference (ACC)}, pp.~4355--4361, IEEE, 2020.

\bibitem{hua2023multi}
M.~Hua, D.~Chen, X.~Qi, K.~Jiang, Z.~E. Liu, Q.~Zhou, and H.~Xu, ``Multi-agent reinforcement learning for connected and automated vehicles control: Recent advancements and future prospects,'' {\em arXiv preprint arXiv:2312.11084}, 2023.

\bibitem{lee2017soft}
C.~Lee, M.~Kim, Y.~J. Kim, N.~Hong, S.~Ryu, H.~J. Kim, and S.~Kim, ``Soft robot review,'' {\em International Journal of Control, Automation and Systems}, vol.~15, pp.~3--15, 2017.

\bibitem{gupta2016learning}
A.~Gupta, C.~Eppner, S.~Levine, and P.~Abbeel, ``Learning dexterous manipulation for a soft robotic hand from human demonstrations,'' in {\em 2016 IEEE/RSJ International Conference on Intelligent Robots and Systems (IROS)}, pp.~3786--3793, IEEE, 2016.

\bibitem{liu2023reinforcement}
X.~Liu, C.~D. Onal, and J.~Fu, ``Reinforcement learning of cpg-regulated locomotion controller for a soft snake robot,'' {\em IEEE Transactions on Robotics}, 2023.

\bibitem{ng1999policy}
A.~Y. Ng, D.~Harada, and S.~Russell, ``Policy invariance under reward transformations: Theory and application to reward shaping,'' in {\em International Conference on Machine Learning}, vol.~99, pp.~278--287, Citeseer, 1999.

\bibitem{ostrovski2017count}
G.~Ostrovski, M.~G. Bellemare, A.~Oord, and R.~Munos, ``Count-based exploration with neural density models,'' in {\em International Conference on Machine Learning}, pp.~2721--2730, PMLR, 2017.

\bibitem{pathak2017curiosity}
D.~Pathak, P.~Agrawal, A.~A. Efros, and T.~Darrell, ``Curiosity-driven exploration by self-supervised prediction,'' in {\em International Conference on Machine Learning}, pp.~2778--2787, PMLR, 2017.

\bibitem{choshen2018dora}
L.~Choshen, L.~Fox, and Y.~Loewenstein, ``Dora the explorer: Directed outreaching reinforcement action-selection,'' {\em arXiv preprint arXiv:1804.04012}, 2018.

\bibitem{badia2020never}
A.~P. Badia, P.~Sprechmann, A.~Vitvitskyi, D.~Guo, B.~Piot, S.~Kapturowski, O.~Tieleman, M.~Arjovsky, A.~Pritzel, A.~Bolt, {\em et~al.}, ``Never give up: Learning directed exploration strategies,'' {\em arXiv preprint arXiv:2002.06038}, 2020.

\bibitem{andrychowicz2017hindsight}
M.~Andrychowicz {\em et~al.}, ``Hindsight experience replay,'' {\em Advances in Neural Information Processing Systems}, vol.~30, 2017.

\bibitem{fang2018dher}
M.~Fang, C.~Zhou, B.~Shi, B.~Gong, J.~Xu, and T.~Zhang, ``Dher: Hindsight experience replay for dynamic goals,'' in {\em International Conference on Learning Representations}, 2018.

\bibitem{ding2019goal}
Y.~Ding, C.~Florensa, P.~Abbeel, and M.~Phielipp, ``Goal-conditioned imitation learning,'' {\em Advances in Neural Information Processing Systems}, vol.~32, 2019.

\bibitem{mnih2015human}
V.~Mnih, A.~K. others, G.~Ostrovski, {\em et~al.}, ``Human-level control through deep reinforcement learning,'' {\em Nature}, vol.~518, no.~7540, pp.~529--533, 2015.

\bibitem{lillicrap2015continuous_long}
T.~P. Lillicrap, J.~J. Hunt, A.~Pritzel, N.~Heess, T.~Erez, Y.~Tassa, D.~Silver, and D.~Wierstra, ``Continuous control with deep reinforcement learning,'' {\em arXiv preprint arXiv:1509.02971}, 2015.

\bibitem{qi2020novel}
X.~Qi, H.~Shi, T.~Pinto, and X.~Tan, ``A novel pneumatic soft snake robot using traveling-wave locomotion in constrained environments,'' {\em IEEE Robotics and Automation Letters}, vol.~5, no.~2, pp.~1610--1617, 2020.

\bibitem{qi2023bioinspired}
X.~Qi, T.~Gao, and X.~Tan, ``Bioinspired 3d-printed snakeskins enable effective serpentine locomotion of a soft robotic snake,'' {\em Soft Robotics}, vol.~10, no.~3, pp.~568--579, 2023.

\bibitem{luo2014theoretical}
M.~Luo, M.~Agheli, and C.~D. Onal, ``Theoretical modeling and experimental analysis of a pressure-operated soft robotic snake,'' {\em Soft Robotics}, vol.~1, no.~2, pp.~136--146, 2014.

\bibitem{hu2009mechanics}
D.~L. Hu, J.~Nirody, T.~Scott, and M.~J. Shelley, ``The mechanics of slithering locomotion,'' {\em Proceedings of the National Academy of Sciences}, vol.~106, no.~25, pp.~10081--10085, 2009.

\bibitem{hu2012slithering}
D.~L. Hu and M.~Shelley, ``Slithering locomotion,'' in {\em Natural locomotion in fluids and on surfaces: swimming, flying, and sliding}, pp.~117--135, Springer, 2012.

\bibitem{schulman2017proximal}
J.~Schulman, F.~Wolski, P.~Dhariwal, A.~Radford, and O.~Klimov, ``Proximal policy optimization algorithms,'' {\em arXiv preprint arXiv:1707.06347}, 2017.

\end{thebibliography}
\end{document}